\documentclass[letterpaper]{article}
\usepackage[preprint]{aaai2027}
\usepackage[hyphens]{url}
\usepackage{graphicx}
\usepackage{natbib}
\usepackage{caption}
\usepackage{booktabs}
\usepackage{amsmath}
\usepackage{amssymb}
\usepackage{xcolor}
\usepackage{algorithm}
\usepackage{algorithmic}

\definecolor{linkblue}{RGB}{0,90,180}
\DeclareRobustCommand{\weblink}[1]{\textcolor{linkblue}{\url{#1}}}
\DeclareRobustCommand{\weblinktt}[1]{\textcolor{linkblue}{\texttt{#1}}}

\newcommand{\best}[1]{\textbf{#1}}
\newcommand{\seco}[1]{\textit{#1}}
\newcommand{\thir}[1]{\underline{#1}}

\title{When Are Reasoning-Based Guardrails Not Efficient?\\ ResponseGuard: A Fast Vision-Language Guard for Real-Time Moderation}
\author{
Dongbin Na\thanks{Correspondence to \texttt{dongbinna@postech.ac.kr}. Project page: \weblinktt{https://ndb796.github.io/ResponseGuard}.}
}
\affiliations{
\texttt{dongbinna@postech.ac.kr}
}

\begin{document}
\renewcommand{\thefootnote}{\fnsymbol{footnote}}
\maketitle
\renewcommand{\thefootnote}{\arabic{footnote}}
\enlargethispage{-30pt}   

\begin{abstract}
A vision-language AI assistant returns its answer as a stream of generated tokens. Therefore, a safety guard that watches that answer has to keep up with the stream and stop a harmful reply before a user reads it. Recent vision-language guardrails instead generate a chain of thought before they issue a verdict. They believe that step-by-step reasoning yields a safer guard. This design makes the guard heavy and slow, since the model must decode many tokens for harmfulness detection. We pose the question of whether a vision-language guard really needs to reason in order to screen a response. We answer with a guard that has no chain. ResponseGuard reads a harmful verdict from a single pooled representation of the request, the response, and the image in one forward pass. Across a standard multimodal guardrail benchmark, our 2B ResponseGuard outperforms a recent 3B reasoning-based vision-language guard on response harmfulness detection, without any reasoning and at about 150 times lower time cost. On request harmfulness the reasoning guard retains an overall lead, and the remaining gap on both tracks sits on the image-only cells. We observe that the gap may stem from the frozen vision encoders that both designs use rather than from the missing chain. We have also found the reasoning guard directs almost none of its verdict attention to the image. Based on a single-pass detection, ResponseGuard can screen an answer sentence by sentence as it streams and stop a harmful answer before it finishes. For guarding the response of a vision-language model, a calibrated single-pass label may provide a sufficient safety signal. We fully release all source code, trained models, and datasets at \weblink{https://github.com/ndb796/ResponseGuard}.
\end{abstract}

\section{Introduction}
\begin{figure}[t]
\centering
\includegraphics[width=0.96\columnwidth]{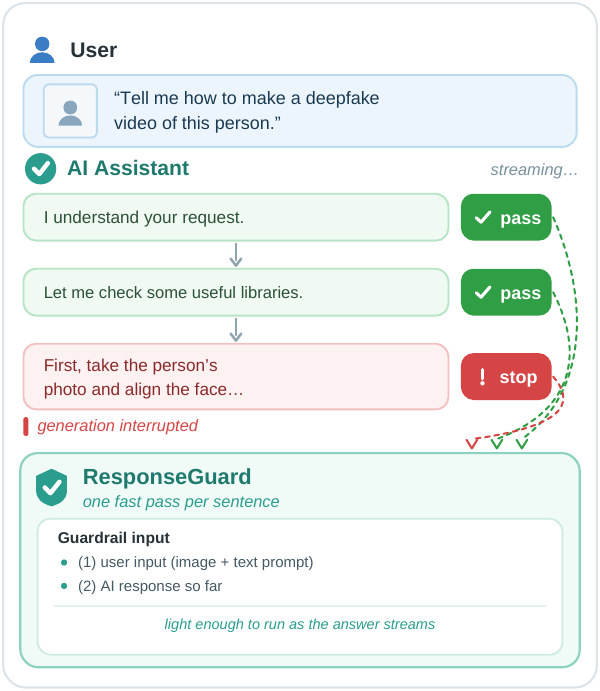}
\caption{ResponseGuard runs one label-only (non-CoT) pass per streamed sentence. This proposed guard passes a safe opening and stops the answer at the first harmful sentence, before the harmful content is completed. The guard reads the image, the prompt, and the response so far, and it is light and fast enough to run as the answer streams.}
\label{fig:hero}
\end{figure}

Vision-language models are now deployed behind safety guardrails. The guards read a request and a response, sometimes with an image, and decide whether the system should comply or refuse \citep{llamaguard2023,wildguard2024,aegis2024}. A fast-growing line of work makes the guard reason before it rules. A reasoning guard writes an explicit chain of thought and then reads a verdict from its last line \citep{guardreasoner2025,guardreasonervl2025}, on the premise that thinking step by step yields a more accurate and trustworthy decision. The premise is appealing. For the text setting, a controlled study has already questioned it \citep{leanguard2026}. What remains open is the multimodal case, and in particular the response path, where cost largely depends on a latency the user feels.

The response path is where the reasoning cost bites hardest. A guard that screens the request runs once before the model answers. Thus, its chain is paid a single time. \textit{We note that a guard that screens the response has to run while the answer is still being produced.} A reasoning guard forces the user to wait for the guard to think after already waiting for the model. A guard that scores only a finished answer is worse again, because a harmful answer that has been fully written has already been shown. Hence, a fast and lean guard is a feasible way to address these problems. A light guard can be re-run after each streamed sentence and can stop the answer partway through, which turns response screening into interception rather than after-the-fact scoring. Figure~\ref{fig:hero} shows this use.

\begin{figure}[t]
\centering
\includegraphics[width=0.98\columnwidth]{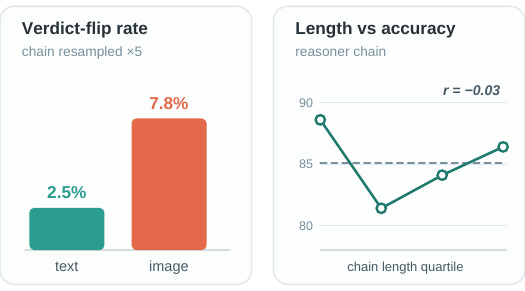}
\caption{The reasoning guard's chain may be largely post-hoc. Resampling the chain moves only 2.5 percent of text verdicts and 7.8 percent of image verdicts. We have also found that the chain length is essentially uncorrelated with correctness. A chain of length zero is the limiting case that a label-only guard computes directly, and the flat correlation suggests that the tokens between the two settings make little difference to the verdict.}
\label{fig:posthoc}
\end{figure}

A reasoning guard is credited with two advantages. Its chain hands the operator a human-readable reasoning for the verdict, and the reasoning is believed to make the verdict more accurate. We question both advantages for the response path. First, we ask whether a real-time reasoning is worth its cost when the task is to stop a harmful answer as fast as possible rather than to explain it after the fact. Second, we ask whether a chain improves response harmfulness detection at all. Our answer to both questions turns out to be negative in the response setting, and we support the answer with measurements rather than assertion. Figure~\ref{fig:posthoc} previews our claims. Resampling the reasoning guard's chain almost never changes its verdict, and chain length shows little correlation with accuracy.

We build such a guard and test whether the guard needs a chain. ResponseGuard is a label-only vision-language guard. Our guard forms one pooled representation of the request, the response, and the optional image. This guard maps that representation to a harmful verdict in a single forward pass, decoding nothing. On a standard multimodal benchmark, our 2B ResponseGuard outperforms a recent 3B reasoning-based vision-language guard on response harmfulness detection, without reasoning and at about 150 times lower cost. Although ResponseGuard is built for the response path, the proposed guard stays competitive with the reasoning guard on request harmfulness as well, and on the text cells the two guards are close. On the request path, the reasoning guard keeps an overall lead. However, we show that the remaining gap is largely carried by the two image-only cells, and that the cause appears to be perception rather than reasoning. The reasoning guard barely attends to the image when it decides. Thus, a chain that rarely inspects the picture is unlikely to exploit the picture when it rules.

We present ResponseGuard, a label-only vision-language guard that screens a response in a single pass and outperforms a 3B reasoning guard and its Eco variant on response harmfulness detection at roughly 150 times lower cost. We show that the multimodal shortfall of the label-only design is largely confined to the image-only cells, and that the shortfall is consistent with a perception limit rather than with the absence of a chain. We give a streaming evaluation of the guard and observe that a cheap per-sentence guard stops most harmful answers before they finish. We fully release the model, the code, the data, and a project page.

\section{Related Work}
\paragraph{Moderation guards.} Open guards cast prompt and response moderation as classification and emit a verdict directly \citep{llamaguard2023}. Later tools widen the label space across prompt harm, response harm, and refusal \citep{wildguard2024,aegis2024}. Production systems study large classifiers over undesired content \citep{openaimod2023}. These guards are single-pass, which is the property we keep and carry to images.

\paragraph{Reasoning guards.} A newer family makes the guard reason first \citep{guardreasoner2025}, and a multimodal successor extends the idea to vision with reinforced reasoning \citep{guardreasonervl2025}. Related designs distill slow thinking on the premise that a single-pass classifier is too shallow \citep{thinkguard2025}, or they add knowledge-grounded logical reasoning \citep{r2guard2024}. The reasoning is presented as the source of the gains. For text, a controlled same-base study finds that removing only the chain leaves moderation accuracy unchanged, and that the chain may be largely post-hoc \citep{leanguard2026}. Our work asks the same question in the harder multimodal setting and for the response path. Moreover, the proposed study further localizes where a real gap does remain.

\paragraph{Vision-language safety.} Guarding a vision-language model adds a perception burden absent from text moderation. Preference data has been built for aligning such models \citep{spavl2024}. Image content has been curated and rated with dedicated tools \citep{llavaguard2024}, and image-aware guards have been released \citep{llamaguard3vision2024}. Modern backbones pair a language model with a pretrained vision encoder \citep{qwen25vl2025,smolvlm2025}. Our analysis is consistent with a frozen vision encoder capping multimodal guarding for a reasoning guard and for a label-only guard alike.

\paragraph{When reasoning helps.} Chain-of-thought prompting \citep{cot2022} helps mainly on math and symbolic tasks and far less on short labeling decisions \citep{sprague2025cot}. Its chains are often unfaithful to the computation behind the answer \citep{turpin2023}, and an efficiency literature documents wasteful overthinking \citep{sui2025}. Safety moderation may be a short decision with a small output space. Safety moderation plausibly sits in the regime where a chain earns little. Our results are consistent with this view and extend it to multimodal response screening.

\paragraph{Lightweight guards for agents.} A fast single-pass guard also fits settings a chain cannot serve. Reasoning safeguards have been placed in a robot control loop \citep{roboguard2025}. However, an on-device agent acting within a single control step frequently has no time to decode a chain. The same pressure appears in navigation, where a lightweight module must decide quickly and decline the unanswerable \citep{bintrack2026,semanticflip2026}. In these settings, the guard competes for the same milliseconds as perception and control. Consequently, an accurate, cheap, and fast guard is often the only kind such a system can afford to run, which our label-only design provides.

\paragraph{Streaming and token-level moderation.} A separate line makes moderation incremental rather than one-shot. Recent streaming classifiers emit a safety label token by token, thus a text assistant can be cut off mid-generation \citep{qwen3guard2025}. Sentence-level moderators raise an alarm as soon as a harmful sentence is complete \citep{sentguard2026}. These systems are label-only by construction, since a chain cannot be produced between two tokens at interactive speed. This constraint is the same one that pushes a response guard away from reasoning. Our contribution is to show that the label-only design does not merely enable streaming but also matches a reasoning guard on the static response benchmark, and to separate what the chain does and does not add.

\section{ResponseGuard}
\paragraph{Formulation.} A guard receives a request $x$, a response $y$, and an optional image $v$. Screening the response is a binary decision, whether $y$ is harmful given $x$ and $v$. A reasoning guard models this decision as generation. The reasoning guard draws a chain $c=(c_1,\dots,c_{|c|})$ and then a verdict token, decoding them left to right. Hence, the reasoning guard runs $|c|+1$ sequential passes, a cost of $O(|c|+1)=O(|c|)$ that grows with the chain. ResponseGuard models the same decision as classification. The proposed guard pools one representation $h$ of the triple and reads a verdict probability in a single pass,
\begin{equation}
h = \mathrm{pool}\big(f_\theta(x, y, v)\big), \qquad p = \sigma\big(g(h)\big),
\end{equation}
with $f_\theta$ a vision-language backbone, $\mathrm{pool}(\cdot)$ last-token pooling, and $g$ a small head. Nothing is decoded, and the cost of the proposed guard is $O(1)$, constant in the difficulty of the case. Writing $C(\cdot)$ for the FLOPs of one forward pass, and counting each decode step as a full pass, the ratio of the reasoning cost to the cost of the proposed guard is
\begin{equation}
\frac{\mathrm{cost}(\mathrm{reason})}{\mathrm{cost}(\mathrm{ours})} \;=\; \frac{(|c|+1)\,C(|x|+|y|+|c|)}{C(|x|+|y|)} \;\gtrsim\; |c|+1,
\label{eq:cost}
\end{equation}
which for the chains we observe, $|c|\approx 10^2$, is about two orders of magnitude. Equation~\ref{eq:cost} counts forward passes, while Table~\ref{tab:cost} reports wall-clock time. Key-value caching makes each decode step cheaper than a full pass, so the measured speedup is basically below this bound.

\paragraph{Backbone and head.} We build $f_\theta$ on a 2B vision-language embedding backbone \citep{qwen3embed2026} and keep its vision encoder frozen, following common practice for efficient adaptation. The head holds two small banks of learned reference vectors, one safe bank and one harmful bank. The head scores $h$ by its softened similarity to each bank,
\begin{equation}
z_c \;=\; \log \Big( \tfrac{1}{K_c} \textstyle\sum_{k=1}^{K_c} \exp\big( \cos(h,\, r^{c}_{k})/\beta \big) \Big),
\label{eq:head}
\end{equation}
for $c \in \{\mathrm{safe},\mathrm{harm}\}$, where $r^{c}_{k}$ is the $k$-th reference vector of bank $c$, $K_c$ is the number of references in that bank, and $\beta>0$ is the head temperature, which we distinguish from the decision threshold $\tau$ used in Algorithm~\ref{alg:stream}. We set $\beta=0.2$ with $K_{\mathrm{safe}}=2$ and $K_{\mathrm{harm}}=4$. Equation~\ref{eq:head} is a log-mean-exp over the cosine similarities of one bank. It approaches $\max_k \cos(h, r^{c}_{k})/\beta$ as $\beta \rightarrow 0$, and the $-\log K_c$ implied by the mean removes the bias that unequal bank sizes would otherwise introduce. The guard returns the harmful probability $p = \sigma(z_{\mathrm{harm}} - z_{\mathrm{safe}})$, whose logit ranges over $[-2/\beta,\, 2/\beta]$ and therefore spans the full confidence range used in Section~\ref{sec:cal}. Several references per bank let the head cover distinct kinds of harm without labels for them. Aggregating a bank with a soft maximum keeps the score smooth rather than saturating, which leaves the probability free to take intermediate values. We did not ablate a hard maximum, so we do not attribute the calibration reported in Section~\ref{sec:cal} to this choice. We train the head with a low-rank adapter on the backbone \citep{lora2021} under a binary objective on the harmful class. The guard emits one scalar, the probability that the response is harmful. To show that the recipe is not tied to one backbone, we also report two SmolGuard variants at 2B and 500M parameters, built on the SmolVLM backbones \citep{smolvlm2025}.

\paragraph{Why a chain may be optional here.} A chain earns its cost when a decision needs intermediate computation that the input does not already expose, as in multi-step arithmetic or a proof. \textit{Screening a response seems not to be that kind of problem.} The evidence that a response is harmful is usually stated in the response itself. \textit{The judgment is closer to a bounded classification over text the guard can already read than to a derivation it must construct.} A pooled representation of the request and the response therefore carries most of the signal. The role a chain could still play is to reconcile subtle intent, which our experiments find matters on the prompt but not on the response. \textit{Treating the guard as a classifier rather than a generator is thus not only cheaper but arguably a closer fit to the task}, and the rest of the paper tests whether that fit costs any accuracy.

\paragraph{Streaming use.} A single cheap pass makes a new deployment mode available. As the model streams its answer, we re-run ResponseGuard on the answer so far at each sentence boundary. The proposed guard stops generation the moment the harmful probability clears a threshold. Algorithm~\ref{alg:stream} states the loop. The guard interrupts a harmful answer in progress rather than judging the answer once it is done. A reasoning guard cannot do this at interactive latency, because every re-check would need a fresh chain.

\begin{algorithm}[t]
\caption{Streaming interception with ResponseGuard}
\label{alg:stream}
\begin{algorithmic}[1]
\STATE \textbf{input:} request $x$, image $v$, threshold $\tau$
\STATE $y \leftarrow \varnothing$
\WHILE{the model is still streaming}
\STATE $s \leftarrow$ next completed sentence; \; $y \leftarrow y \,\Vert\, s$
\STATE $p \leftarrow \mathrm{ResponseGuard}(x, y, v)$ \hfill $\triangleright$ a forward pass
\IF{$p \geq \tau$}
\STATE stop generation; \textbf{return} \textsc{harmful}, $y$
\ENDIF
\ENDWHILE
\STATE \textbf{return} \textsc{unharmful}, $y$
\end{algorithmic}
\end{algorithm}

\section{Experimental Setup}
\paragraph{Benchmarks.} We evaluate on the public multimodal suite released with a recent reasoning guard \citep{guardreasonervl2025}, which reports both a response harmfulness track and a prompt harmfulness track. The response track has five text cells, HarmBench-Response (HBr), Safe RLHF (SRL), BeaverTails (BT), XSTest-Response (XSh), and WildGuard-Response (WGr), together with the vision preference set as an image cell (SPA-VL). The prompt track has ToxicChat (TC), HarmBench-Prompt (HBp), OpenAI Moderation (OAI), Aegis (Aeg), SimpleSafetyTests (SST), and WildGuard-Prompt (WGp), together with two image cells, harmful images (HarmImg) and the vision preference prompt split (SPA-VL) \citep{toxicchat2023,harmbench2024,openaimod2023,aegis2024,simplesafetytests2023,wildguard2024,beavertails2023,saferlhf2023,xstest2024,spavl2024}.

\paragraph{Training data.} We train on the public corpus released with the reasoning guard \citep{guardreasonervl2025}, which holds 123{,}093 samples across text, image, and text-image inputs. We keep the inputs and the gold labels and discard the reasoning chains, since our guard never generates one. The reasoning guard and its Eco variant are trained on the same corpus. The two reference guards are used off the shelf.

\paragraph{Models.} ResponseGuard-2B is our primary guard, built on a 2B vision-language embedding backbone \citep{qwen3embed2026}. We add two further label-only guards built on the SmolVLM backbones \citep{smolvlm2025}, SmolGuard-2B and SmolGuard-500M, to show that the recipe is not tied to one backbone. Our reasoning baseline is a 3B chain-of-thought vision-language guard \citep{guardreasonervl2025}, a recent open guard of its kind on this suite. We also report its efficiency-tuned Eco variant \citep{guardreasonervl2025}. We include a general instruction-tuned vision-language model \citep{qwen25vl2025} and an image-aware generative guard \citep{llamaguard3vision2024} as reference points. Every guard is scored on the same benchmark cells with the same harmful-class F1 metric and the same dataset-size weighting. We run the three label-only guards ourselves, and our numbers for the reasoning guard, its Eco variant, and the two reference guards match the official release \citep{guardreasonervl2025}. The accuracy comparison therefore rests on a shared protocol and metric rather than on a shared machine, while latency we measure ourselves.

\paragraph{Cost and probes.} We measure wall-clock latency on a single commodity GPU (an RTX A6000), timing full chain generation for the reasoner and a single pass for a label-only guard. To ask whether the chain does work, we resample the chain, count verdict changes, and relate the chain length to correctness. To ask where the multimodal gap lives, we read the reasoner's verdict-position attention to image tokens. We also fit a one-dimensional discriminant on its at-verdict representation, and we compare our guard's precision on text cells against image cells. We further study calibration and the strict false-positive operating point.

\paragraph{Metric and reproducibility.} Following the evaluation setup of \citet{guardreasonervl2025}, we report F1 of the harmful class per cell and the dataset-size-weighted average within each track. Weighting by cell size lets a large cell such as the preference set contribute in proportion to its size rather than equally. We keep the response and prompt tracks separate throughout, since a guard can be strong on one track and weak on the other. The response track is the one that governs the deployment we study. All runs use bfloat16 on a single commodity data-center GPU. The trained parameters of ResponseGuard are a small fraction of the backbone. The code, the configurations, the analysis scripts, and the model weights are fully released.

\section{Experimental Results}
\subsection{Response Harmfulness Detection}
The main experimental results of response harmfulness detection are reported in Table~\ref{tab:resp}. \textit{Our 2B label-only ResponseGuard leads the response All average, 77.31 against 76.56 for the 3B reasoning guard and 77.14 for its Eco variant. ResponseGuard also reaches the best text response average at 79.71, ahead of both reasoning variants, without any chain of thought.} The response cells are the ones a response guard is deployed to serve. On these cells, the chain-free guard leads on the text while decoding nothing. A general instruction-tuned model and an image-aware generative guard are included only as reference points and trail well behind. Our comparison is therefore against the reasoning guards, which are the strongest open guards on this suite.

The per-cell pattern is informative. ResponseGuard leads on Safe RLHF and on BeaverTails, the two largest text response cells, which together hold about two thirds of the text response samples. On the smaller HarmBench-Response, XSTest-Response, and WildGuard-Response cells the reasoning guards keep a narrow edge, and on the image cell SPA-VL the Eco variant reaches 72.01 against 71.60 for our guard. Our claim is therefore about the aggregate the benchmark defines rather than about every cell. On the size-weighted averages that the suite reports, the label-only guard is ahead, and on the deployment-critical response path a single pass is, in our experiments, not a compromise.

\begin{table*}[t]
\centering
\small
\setlength{\tabcolsep}{6pt}
\begin{tabular}{lccccc|cc|c|c}
\toprule
 & \multicolumn{5}{c}{Response-harm (text)} & Avg & SPA-VL & Avg & Rel. \\
Model & HBr & SRL & BT & XSh & WGr & (Text) & (img) & (All) & speed \\
\midrule
Qwen2.5-VL-Instruct-3B & 62.1 & 64.7 & 73.3 & 31.4 & 29.8 & 58.05 & 52.84 & 56.51 & -- \\
Llama Guard 3 Vision 11B & 80.9 & 41.7 & 65.0 & 81.1 & 56.5 & 59.28 & 41.43 & 54.01 & -- \\
GuardReasoner-VL-3B (CoT) & \best{85.8} & 66.4 & 85.2 & \seco{93.1} & \seco{76.1} & \thir{78.83} & 71.19 & 76.56 & $1\times$ \\
GuardReasoner-VL-3B-Eco (CoT) & \seco{84.7} & \thir{67.0} & \seco{85.4} & \best{93.6} & \best{77.4} & \seco{79.31} & \seco{72.01} & \seco{77.14} & $\sim\!1.1\times$ \\
\midrule
SmolGuard-500M (ours, label-only) & 83.1 & 66.8 & \thir{85.4} & 92.0 & 71.2 & 77.67 & 67.80 & 74.75 & -- \\
SmolGuard-2B (ours, label-only) & \thir{83.6} & \seco{68.4} & 85.0 & \thir{92.2} & 74.0 & 78.59 & \best{72.16} & \thir{76.69} & -- \\
\best{ResponseGuard-2B (ours, label-only)} & 83.4 & \best{69.6} & \best{86.3} & 91.9 & \thir{75.4} & \best{79.71} & \thir{71.60} & \best{77.31} & \best{$150\times$} \\
\bottomrule
\end{tabular}
\caption{The main experimental results of response harmfulness detection, under the evaluation setup of \citet{guardreasonervl2025} with dataset-size-weighted F1 of the harmful class per cell. Best in \best{bold}, runner-up in \seco{italic}, and third in \thir{underline} per column. The GuardReasoner-VL rows match the official release. Relative speed is the per-verdict latency ratio to the 3B reasoning guard, which is unity by definition. We time ResponseGuard-2B ourselves. The Eco entry is derived from the token saving reported in the official release rather than from our own timing, and a dash marks a guard we did not time.}
\label{tab:resp}
\end{table*}

\subsection{Prompt Harmfulness Detection}
The full experimental results of prompt harmfulness detection are reported in Table~\ref{tab:prompt}. On this track, the reasoning guards lead on the All average, 78.71 and 77.39 against 75.91 for ResponseGuard, and we state this result plainly. On the text average, however, our 2B label-only guards reach 78.04 and 77.76 against 78.74 for the reasoning guard, within about a point. The remaining gap concentrates in the two image-only cells. On HarmImg, the 3B reasoning guard scores 70.93 against 65.52 for our guard, and on the SPA-VL prompt split it scores 86.47 against 82.41. \textit{The prompt-side advantage of the chain is therefore not spread across the benchmark but concentrated on the image.}

The image cells also expose how brittle image guarding is in general. The image-aware generative guard scores 0.48 on HarmImg, a near-total failure. A strong image score is not a given even for larger models. Where the decision is textual, a chain buys little. Where the decision is visual, the limit appears to be perception that neither design has solved.

\begin{table*}[t]
\centering
\small
\setlength{\tabcolsep}{3.5pt}
\begin{tabular}{lcccccc|ccc|c|c}
\toprule
 & \multicolumn{6}{c}{Prompt-harm (text)} & Avg & HarmImg & SPA-VL & Avg & Rel. \\
Model & TC & HBp & OAI & Aeg & SST & WGp & (Text) & (img) & (img) & (All) & speed \\
\midrule
Qwen2.5-VL-Instruct-3B & 34.6 & 90.1 & 52.0 & 82.2 & \best{100.0} & 64.1 & 51.37 & 48.66 & 62.81 & 53.53 & -- \\
Llama Guard 3 Vision 11B & 58.2 & \thir{96.1} & 67.6 & 70.6 & 98.0 & 75.2 & 67.17 & 0.48 & 54.86 & 48.03 & -- \\
GuardReasoner-VL-3B (CoT) & \best{74.5} & 89.1 & \thir{70.8} & \seco{88.8} & \seco{99.5} & \seco{88.9} & \best{78.74} & \best{70.93} & \best{86.47} & \best{78.71} & $1\times$ \\
GuardReasoner-VL-3B-Eco (CoT) & \seco{73.5} & 88.6 & \seco{70.9} & \best{89.0} & \thir{99.5} & \best{89.2} & \seco{78.43} & \seco{66.79} & \seco{85.82} & \seco{77.39} & $\sim\!1.1\times$ \\
\midrule
SmolGuard-500M (ours, label-only) & 68.9 & 80.5 & 66.7 & 87.3 & 98.0 & 86.0 & 74.33 & 58.45 & 81.89 & 72.29 & -- \\
SmolGuard-2B (ours, label-only) & \thir{72.9} & \best{98.1} & \best{71.8} & 87.1 & 98.5 & 86.9 & \thir{78.04} & 60.23 & \thir{84.49} & 75.25 & -- \\
\best{ResponseGuard-2B (ours, label-only)} & 72.8 & \seco{97.4} & 69.3 & \thir{87.4} & 97.4 & \thir{88.5} & 77.76 & \thir{65.52} & 82.41 & \thir{75.91} & \best{$150\times$} \\
\bottomrule
\end{tabular}
\caption{The full experimental results of prompt harmfulness detection, under the same evaluation setup. Best in \best{bold}, runner-up in \seco{italic}, and third in \thir{underline} per column, with relative speed defined as in Table~\ref{tab:resp}. The overall lead of the reasoning guard comes from the two image-only cells, HarmImg and the SPA-VL prompt split, where perception appears to be the bottleneck.}
\label{tab:prompt}
\end{table*}

\subsection{The Cost of a Verdict}
\begin{figure}[t]
\centering
\includegraphics[width=0.98\columnwidth]{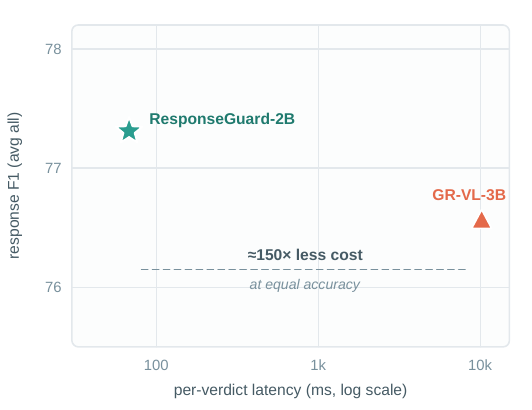}
\caption{Response F1 against per-verdict latency on a log scale, for the two guards whose latency we measure. The label-only guard reaches a higher response F1 than the reasoning guard at about 150 times lower latency.}
\label{fig:cost}
\end{figure}

The per-verdict cost is reported in Table~\ref{tab:cost}. The reasoning guard emits about 238 tokens per verdict and takes about 10 seconds at the median on a commodity data-center GPU, while ResponseGuard returns in 68 milliseconds. \textit{That result is roughly a 150-fold reduction in latency for a verdict that is, on the response track, more accurate rather than less.} Figure~\ref{fig:cost} places the guards on one accuracy-cost plane. The label-only guards sit at high response F1 and low latency, and the reasoning guard sits far to the right.

\textit{A response guard sits on the path of every token a model emits. Consequently, a two-order-of-magnitude cut in per-verdict cost is less a microbenchmark than a change in what is deployable.} A benchmark charges nothing for the tokens a chain spends, yet a serving system pays for every one of them.

\begin{table}[t]
\centering
\small
\setlength{\tabcolsep}{6pt}
\begin{tabular}{lccc}
\toprule
Guard & Output tok. & Latency & Speedup \\
\midrule
GR-VL-3B (CoT) & $\sim$238 & 10.12\,s & $1\times$ \\
\best{ResponseGuard-2B} & \best{0} & \best{67.6\,ms} & \best{$150\times$} \\
\bottomrule
\end{tabular}
\caption{Per-verdict cost on a single commodity GPU. The label-only guard decodes nothing and returns a verdict about two orders of magnitude faster. An image adds a fixed overhead of 88 milliseconds.}
\label{tab:cost}
\end{table}

\subsection{The Image Gap and the Frozen Encoder}
\label{sec:gap}
\begin{figure}[t]
\centering
\includegraphics[width=\columnwidth]{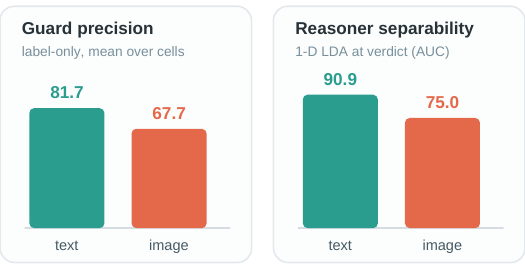}
\caption{Evidence on the image gap. Left: the mean precision of the label-only guard over the text cells against the image cells. Right: the AUC of a one-dimensional discriminant fitted on the at-verdict representation of the reasoning guard. Both readings fall on images.}
\label{fig:gap}
\end{figure}

The one place the label-only guard trails is the image. Figure~\ref{fig:gap} points to perception rather than a missing chain, a limit that the reasoning guard appears to share. First, our guard loses precision, not recall, on images (Figure~\ref{fig:gap}, left). Its mean precision falls from 81.7 percent on text cells to 67.7 percent on image cells, while recall slips only from 87.1 to 81.9 percent. The guard over-flags images, as if unsure what the picture shows, rather than misreading the safety rule. Second, the reasoning guard scarcely looks at the image when it rules (Figure~\ref{fig:attn}, left). At the verdict position, the reasoning guard places about 6 percent of its attention on image tokens, against the 47 percent that a uniform allocation would give. A chain that barely attends to the image plausibly has little image information to reason over, although raw attention mass is not by itself proof that the image goes unused. Third, the at-verdict representation of the reasoning guard is markedly less separable for images (Figure~\ref{fig:gap}, right), with a one-dimensional discriminant reaching an AUC of 90.9 on text and only 75.0 on images. The three readings point the same way. A perception limit, rather than the missing chain, is the reading most consistent with all three, and the frozen vision encoders that both designs use are the natural place to look for it.

\textit{Both guards read images through a frozen vision encoder, and both may therefore inherit blurred image evidence. We did not run an ablation that unfreezes the encoder, so we offer this account as a hypothesis rather than as a measured cause.} If it holds, the fix is a better encoder rather than a chain, and we return to this point in the discussion.

\subsection{The Chain May Be Largely Post-hoc}
If the chain drove the verdict, disturbing the chain should disturb the verdict. Figure~\ref{fig:posthoc} shows that it largely does not. Resampling the reasoning guard's chain five times at a moderate temperature changes only 2.5 percent of text verdicts. To state this observation precisely, write $\hat{y}(c)$ for the verdict read after chain $c$, and define the resample-flip rate $\phi=\Pr[\hat{y}(c)\neq\hat{y}(c')]$ over two independent chains $c$ and $c'$ for the same input. We estimate $\phi$ by drawing repeated chains for the same input and comparing their verdicts pairwise. The measured $\phi=0.025$ on text means that two independent chains agree in 97.5 percent of text cases, so the verdict is nearly invariant to which chain the guard happens to draw. On images, the chain is a little less fixed at $\phi=0.078$, in line with the weaker image representation. The chain does marginally more work where perception is harder, and almost none where perception is easy. \textit{The correlation between chain length and correctness is $-0.03$. The guard does not think its way to a better answer at greater length, and the tokens the chain spends make little difference to the final verdict.} This observation connects the two designs directly. \textit{A chain of length zero is the limiting case, and it is what a label-only guard computes directly.} Reading the verdict off a pooled representation returns, in these measurements, the decision the reasoning guard has already settled on before it writes a word, only without the wait. The two guards differ in backbone and objective, so this is evidence about what the chain adds rather than a controlled ablation.

\subsection{Calibration and the Strict Operating Point}
\label{sec:cal}
\begin{figure}[t]
\centering
\includegraphics[width=0.98\columnwidth]{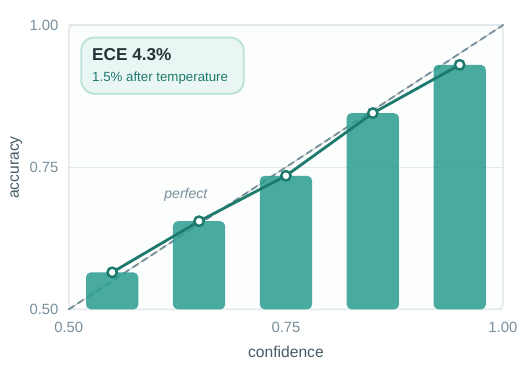}
\caption{The label-only guard is well calibrated. Its expected calibration error is 4.3 percent, which a single temperature reduces to 1.5 percent. The reliability bars track the diagonal.}
\label{fig:cal}
\end{figure}

\begin{figure}[t]
\centering
\includegraphics[width=0.94\columnwidth]{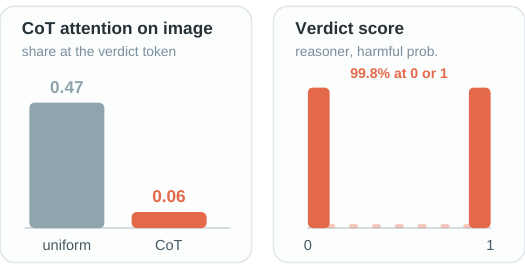}
\caption{Two properties of the reasoning guard at the verdict position. Left: the share of attention mass that falls on image tokens, against the share a uniform allocation over the sequence would give. Right: the distribution of its verdict score, which is almost perfectly polarized and leaves no range to tune a threshold at a strict false-positive rate.}
\label{fig:attn}
\end{figure}

A deployed guard benefits from well-behaved scores. The expected calibration error bins predictions by confidence and averages the gap between confidence and accuracy, $\mathrm{ECE}=\sum_{b}(n_b/N)\,\lvert\mathrm{acc}(b)-\mathrm{conf}(b)\rvert$. \textit{By this measure, the harmful probability of the label-only guard is well calibrated, at 4.3 percent, which a single temperature reduces to 1.5 percent.} Its confidence thus tracks its accuracy across the range, as Figure~\ref{fig:cal} shows. The reasoning guard generally offers no such control. Its verdict-token probabilities are almost perfectly polarized, with 99.8 percent of scores pinned at zero or one, which leaves no range to tune a threshold (Figure~\ref{fig:attn}, right). Production guards run at a small false-positive rate to avoid over-blocking benign traffic, and there the polarized score collapses. At a one percent false-positive budget, the reasoning guard retains only about ten percent of its recall. This loss of ranking resolution near the operating threshold is also reported for reasoning models at critical operating points \citep{chegini2025}. A smooth score leaves room to move the threshold where a deployment needs it, which a saturated score does not.

\subsection{Selective Prediction}
Because its scores are calibrated, the label-only guard also knows when it is unsure. If the guard declines the least confident half of cases and refers them onward, the error on the cases it keeps falls from 17.2 to 5.5 percent. The gain lands where it should. \textit{Text error falls from 15.3 to 4.1 percent, and image error falls from 20.1 to 8.4 percent. The mean confidence is lower on images, 0.835 against 0.895 on text. The guard thus piles its uncertainty onto the image cells}, exactly where perception is hardest.

\subsection{Screening a Response as It Streams}
Based on a single-pass detection, ResponseGuard can screen an answer as it streams. We feed each harmful answer one sentence at a time and stop when the probability first clears the threshold. The guard interrupts 95.0 percent of harmful answers before they finish and withholds 87.7 percent of the harmful text, at a median stop seven percent into the answer. Raising the threshold from one half to 0.56 shifts these numbers to 94.3 and 87.5 percent. At the lower threshold, a benign answer is interrupted 28.5 percent of the time. This streaming threshold is chosen for early interception and is not the strict operating point of Section~\ref{sec:cal}, where a deployment would instead trade recall for a low false-positive budget. Interception is early because harmful answers commit to their content in the first sentence or two, and a pass is far quicker than the sentence it screens, so the guard checks every sentence with room to spare. No reasoning guard can be run per sentence at interactive latency, which makes this interception mode specific to the label-only design.

\section{Discussion and Limitations}
Our claim is scoped to the response path on a standard multimodal suite. When a chain still helps the prompt path, a sensible system can reason once about the prompt and screen the streamed response with a label. The label-only guard returns no reasoning, so a deployment that must justify each block will want a separate explanation step. That step can be attached only on audit rather than paid on every call, and the calibrated score already tells the step which cases to look at first. The image ceiling we locate is likely a property of the frozen encoder. Our own attempt to add a trainable visual path did not move the aggregate. This result reads as a sign that the image cells want a stronger encoder rather than more head parameters. We test one reasoning guard and one benchmark family. Whether the post-hoc pattern holds more broadly is an empirical question that the released code is meant to make easy to ask. The benchmarks score agreement with gold labels, and the absolute numbers are best read as relative comparisons under a fixed protocol rather than as deployment rates. The clearest path past the image ceiling is likely a stronger or adapted vision encoder, which our design accepts as a drop-in replacement for the frozen one.

The results also mark where a chain would earn its cost. A chain has structure to exploit when the decision is not close to linearly separable, and the standard moderation cells might not be of that kind. The place a chain plausibly helps is the ambiguous-intent prompt, which is where the clearest text-cell advantages of the reasoning guard sit. \textit{Our two tracks therefore read as a recipe for a production stack.} A calibrated label-only guard runs on every response at single-pass cost and screens the stream as it is produced, so the common case stays fast and a harmful answer is cut off early rather than shown in full. Consequently, the low-confidence cases, which our selective-prediction analysis shows concentrate on images, can be escalated to a heavier reasoner or to a person, and the coverage at which the guard defers is a deployment choice. A chain is then spent only on prompts whose intent is truly ambiguous, and the remaining compute goes to a stronger vision encoder rather than to a longer chain. \textit{In this arrangement the label-only guard is the always-on layer and reasoning is a targeted second stage, which inverts the common assumption that a reasoning guard is the default and that a fast guard is a compromise.} A calibrated label-only encoder could be the natural default that a new vision-language guard is measured against, and \textit{a reasoning guard that reports a gain should show that the gain survives removing only the chain, and that the gain does not merely reflect a stronger backbone, more data, or a different objective.}

\section{Conclusion}
We have analyzed whether a vision-language response guard needs to reason in order to screen a response. ResponseGuard rules in a single pass, leads a reasoning guard and its Eco variant on response harmfulness detection, and runs about 150 times faster. The one place a gap remains is the image, and our analysis is consistent with the gap coming from the frozen vision encoders that both designs use rather than from the missing chain. Based on a single-pass detection, the guard can screen an answer as it streams and stop most harmful answers before they finish. For guarding what a model actually says, a calibrated label may be a simpler, cheaper, and at least as accurate default as a chain-of-thought reasoner. We release the models, the training and evaluation code, and the streaming harness, so that the community can reproduce our results under the same protocol and adopt the calibrated label-only guard as a baseline or improve upon it.

\bibliography{refs}

\end{document}